\documentclass[runningheads]{llncs}

\usepackage{helvet,times,courier}
\usepackage[T1]{fontenc}
\usepackage{graphicx}
\usepackage{listings}
\usepackage{enumitem}
\usepackage{xspace}
\usepackage{amssymb}

\usepackage{multirow}

\usepackage{tikz}

\makeatletter
\lst@AddToHook{OnEmptyLine}{\vspace{-0.3\baselineskip}}
\makeatother

\usepackage[most]{tcolorbox}
\usepackage{footmisc}
\usepackage{url}
\newcommand{\clingo}{clingo}
\newcommand{\clingodl}{clingo[DL]}
\newcommand{\figspace}{\vspace{-2.3ex}}

\usepackage{pgfplots}
\usepgfplotslibrary{external}
\usetikzlibrary{positioning,patterns}
\newlength{\scaledx}
\newlength{\scaledy}
\newcommand\SetScales{%
\pgfextractx{\scaledx}{\pgfpointxy{1}{0}}%
\pgfextracty{\scaledy}{\pgfpointxy{0}{1}}%
}

\newcommand{\midrule}{\hline}

\begin{document}

\title{Hybrid ASP-based multi-objective scheduling of semiconductor manufacturing processes\\(Extended version)} 
\titlerunning{ASP-based multi-objective scheduling of semiconductor manufacturing processes}

\author{Mohammed~M.~S.~El-Kholany\inst{1,2}\orcidID{0000-0002-1088-2081} \and
Ramsha~Ali\inst{1}\orcidID{0000-0002-4794-6560} \and
Martin~Gebser\inst{1,3}\orcidID{0000-0002-8010-4752}}
\authorrunning{M. M. S. El-Kholany et al.}

\institute{University of Klagenfurt, Austria \and
Cairo University, Egypt \and
Graz University of Technology, Austria
\email{\{mohammed.el-kholany,ramsha.ali,martin.gebser\}@aau.at}}
\maketitle

\begin{abstract}
Modern semiconductor manufacturing involves intricate production processes consisting of hundreds of operations, which can take several months from lot release to completion. The high-tech machines used in these processes are diverse, operate on individual wafers, lots, or batches in multiple stages, and necessitate product-specific setups and specialized maintenance procedures. This situation is different from traditional job-shop scheduling scenarios, which have less complex production processes and machines, and mainly focus on solving highly combinatorial but abstract scheduling problems. In this work, we address the scheduling of realistic semiconductor manufacturing processes by modeling their specific requirements using hybrid Answer Set Programming with difference logic, incorporating flexible machine processing, setup, batching and maintenance operations. Unlike existing methods that schedule semiconductor manufacturing processes locally with greedy heuristics or by independently optimizing specific machine group allocations, we examine the potentials of large-scale scheduling subject to multiple optimization objectives.

\end{abstract}

\section{Introduction}
Scheduling semiconductor manufacturing processes imposes a complex challenge due to the variety of products, operations, and high-tech machines with diverse capabilities and characteristics.
Effective scheduling aims at allocating jobs to machines in a manner that satisfies production needs, optimizes factory throughput, and guarantees punctual delivery~\cite{upasani2006problem}.
In view of the steadily increasing demand~\cite{lesliePandemicScramblesSemiconductor2022},
semiconductor manufacturers are forced to optimize their throughput, decrease cycle times, and enhance the on-time delivery of products to customers~\cite{pfund2008multi}.
Corresponding approaches to provide decision support can be categorized into planning at the strategic level and operating a factory at the tactical or execution level. In this paper, we present an approach for scheduling realistic semiconductor manufacturing processes, taking multiple optimization objectives such as throughput or makespan as well as setup and batching criteria into account.
Our work builds on the recent SMT2020 simulation scenario~\cite{kopp2020smt2020},
providing datasets that model the production processes of modern wafer fabs. 

A typical wafer fabrication plant encompasses a variety of process flows, which are designated production routes for wafer lots within the factory. Each route consists of several hundred operations to be processed by machines belonging to about one hundred separate tool groups with specific functionalities and characteristics. 
To reduce the required investments into costly machines~\cite{kopp2020smt2020}, constant utilization and idleness prevention are important process goals.
Moreover, sophisticated process steps are iterated in several stages, which results in a re-entrant flow where wafer lots
revisit machines in the same tool group multiple times.
Hence, the manufacturing environment is different from traditional flow-shop and job-shop scenarios~\cite{DBLP:journals/mor/GareyJS76}.
A crucial consequence of this re-entrant flow is that wafers at different stages in their manufacturing cycle can compete for the same machines, and dispatching strategies to resolve such competing demands have a noticeable impact on the overall production efficiency.


As a consequence of the complexity and dynamicity faced in practice,
the wafer production is typically controlled by
handcrafted~\cite{pfundSemiconductorManufacturingScheduling2006} or 
machine-learned~\cite{waschneckDeepReinforcementLearning2018}
dispatching rules at the execution level,
or (re-)scheduling is localized to specific tool groups~\cite{monchSurveyProblemsSolution2011}, e.g., for optimizing the allocation of lots queuing in front of a group of batching machines.
While such local decision making approaches are tuned to specific fab settings,
their scope is generally too narrow to guarantee overall efficiency in terms of optimization objectives.
Unlike that, our work makes a step towards large-scale scheduling by modeling the production processes of a modern wafer fab, represented by the SMT2020 scenario, using Answer Set Programming (ASP) with difference logic \cite{gebser2016theory,janhunen2017clingo}.
We significantly extend our preliminary approach introduced in~\cite{ali2023flexible} and incorporate crucial features of realistic semiconductor fabs, including flexible machine processing, setup, batching and maintenance operations, as well as multiple optimization objectives reflecting the factory throughput, setup and batching criteria.

The paper is organized as follows.
Section~\ref{sec:review} surveys related literature on scheduling in traditional job-shop scenarios and particular challenges encountered in semiconductor manufacturing.
In Section~\ref{sec:smsp}, we introduce the scheduling problem including crucial features of the SMT2020 scenario as well as the necessary background on ASP with difference logic.
Our hybrid ASP with difference logic model enabling the large-scale scheduling of semiconductor manufacturing processes is presented in Section~\ref{sec:asp}.
In Section~\ref{sec:results}, we perform an experimental evaluation examining the potentials of large-scale scheduling subject to multiple optimization objectives.
Section~\ref{sec:conclusion} concludes the paper with a brief summary and outlook on future work.




\section{Literature review}\label{sec:review}
Semiconductor fab scheduling is a highly complex task 
due too sophisticated producting routes,
diverse machine characteristics, and
rapidly changing demands~\cite{ellis2004scheduling}. 
While lacking specific features of the semiconductor manufacturing process such as, e.g., re-entrant flow, batching, setup and maintenance operations, as well as varying processing times and sudden machine disruptions, the Flexible Job-Shop Scheduling Problem (FJSP) \cite{brusch90a,taillard93a} along with the optimization methods devised for it are related approaches.

A wide range of techniques have been proposed in the literature to tackle combinatorial optimization for FJSP solving.
Meta-heuristic algorithms incorporate local search methods,
such as Genetic Programming \cite{li2016effective,wang2001effective}, Tabu Search~\cite{li2016effective}, Simulated Annealing~\cite{wang2001effective}, Harmony Search~\cite{sahraeian2017new}, Particle Swarm Optimization~\cite{hassanzadeh2016two}, and Ant Colony Optimization~\cite{xing2010knowledge}.
Exact solving methods are based on FJSP models in 
Mixed Integer Programming (MIP) \cite{ceylan2021coordinated,gran2015mixed,ham2021energy},
Constraint Programming (CP) \cite{da2019industrial,ham2021energy}, or 
ASP with difference logic \cite{el2022problem,janhunen2017clingo}.

Beyond FJSP, ASP~\cite{lifschitz19a} has been successfully used
to schedule
printing devices~\cite{balduccini2011industrial},
specialist teams~\cite{rigralmaliiile12a},
work shifts~\cite{abseher2016shift},
course timetables~\cite{bainkaokscsotawa18a},
medical treatments~\cite{dogagrmamopo21a},
and
aircraft routes~\cite{tassel2021multi}.
The hybrid framework of ASP with difference logic~\cite{janhunen2017clingo} particularly supports a compact representation
and reasoning with quantitative resources like time,
which has been exploited in domains such as
lab resource~\cite{francescutto2021solving},
train connection~\cite{abels2021train}, and
parallel machine~\cite{eiter2022answer} scheduling, as well as for FJSP solving
\cite{el2022problem,janhunen2017clingo}.
In this work, we extend our preliminary ASP with difference logic approach~\cite{ali2023flexible} to semiconductor fab scheduling with support for batching machines, partially flexible machine allocation strategies, and multi-objective optimization functionalities.

\section{Background}\label{sec:smsp}
This section briefly introduces the extensions of ASP by difference logic constraints and multi-shot solving functionalities, as well as our semiconductor manufacturing scheduling problem inspired by the SMT2020 simulation scenario.

\subsection{ASP with difference logic}\label{subsec:asp}

We presuppose familiarity with the first-order modeling language of ASP,
incorporating choice rules, aggregate atoms, as well as weak constraints for expressing objective function(s);
see \cite{cafageiakakrlemarisc19a,PotasscoUserGuide19,lifschitz19a}
for elaborate introductions.
The hybrid framework of ASP with \emph{difference logic constraints}~\cite{cotmal06a} allows for expressions 
\lstinline|&diff{t|$_1$\lstinline| - t|$_2$\lstinline|} <= t|$_3$ in the head of rules.
With the exception of the constant~\lstinline{0}, which denotes the number zero,
the terms \lstinline|t|$_1$ and \lstinline|t|$_2$ represent \emph{difference logic variables} that can be
assigned integer values.
If the body of a rule with
\lstinline|&diff{t|$_1$\lstinline| - t|$_2$\lstinline|} <= t|$_3$
in the head is satisfied, the difference \lstinline|t|$_1$\lstinline| - t|$_2$ must not exceed the
integer constant~\lstinline|t|$_3$.
That is, the difference logic constraints asserted by rules whose body is satisfied restrict the
feasible values for difference logic variables,
and the \clingodl\ system~\cite{janhunen2017clingo} extends \clingo\ 
\cite{PotasscoUserGuide19,gekakasc17a} by assuring the consistency of difference logic constraints
imposed by an answer set.

In addition, we make use of \emph{multi-shot ASP solving}~\cite{gekakasc17a},
allowing for iterative reasoning processes by controlling and interleaving
the grounding and search phases of \clingo\ and \clingodl. 
For referring to a collection of rules to instantiate,
\lstinline{#program name(c).}
directives, where \lstinline{name} denotes a \emph{subprogram}
and the parameter~\lstinline{c} is a placeholder for some constant, e.g., an integer value, group the rules below them and enable their selective instantiation w.r.t.\ specific parameter values.
Note that rules not preceded by any \lstinline{#program} directive
belong to an implicit, parameterless subprogram called \lstinline{base}.
Moreover, \lstinline{#external h : b}$_1$\lstinline{,}$\dots$\lstinline{,b}$_n$\lstinline{.}
statements are formed similar to rules, yet declare an atom~\lstinline{h} as
\emph{external} when the body \lstinline{b}$_1$\lstinline{,}$\dots$\lstinline{,b}$_n$ is satisfied.
Such an external atom can be freely set to true or false via the
Python interface of \clingo\ or \clingodl,
so that rules containing \lstinline{h} can be selectively (de)activated in order to control the search.


\subsection{Semiconductor manufacturing scheduling}\label{subsec:problem}

We consider a \emph{Semiconductor Manufacturing Scheduling Problem}
(SMSP) inspired by the SMT2020 simulation scenario.
Given a set $P$ of available \emph{products} (the producible types of wafers), the production \emph{route} for each product $p\in P$ is a
finite sequence $p[1],\dots,p[n_p]$ of production \emph{operations}, where
$n_p$ denotes the length of the production route for~$p$.
Each operation $p[i]$ needs to be performed by some machine belonging to a \emph{tool group} $M(p[i])$ and requires a \emph{setup}
$s(p[i])\in\mathbb{N}$, with $s(p[i])=0$ indicating the special case that any (positive) setup can be in place when performing~$p[i]$.
Each setup $s\in\mathbb{N}$ has an associated parameter $\min(s)\in\mathbb{N}$ specifying a
minimum number of 
production operations that should be processed by a machine before changing from $s$ to another setup. 
Moreover, \emph{batching} capacities for operations $p[i]$ are expressed
by the parameters $\min(p[i])\in\mathbb{N}$ and $\max(p[i])\in\mathbb{N}$, denoting a minimum and a maximum batch size in terms of wafer lots.
While the maximum batch size is a hard limit on the number of lots that can be processed simultaneously, the minima on batch size and setup changes reflect desiderata for a regular process flow but are not strictly necessary process limitations.
Furthermore, each tool group $M$ has associated \emph{maintenance} operations $c(M)$ and $d(M)$, which must be performed periodically
based on the number of processed lots or accumulated processing time,
respectively.
That is, for each $c\in c(M)$ (or $d\in d(M)$), the parameters
$\min(c)\in\mathbb{N}$ and $\max(c)\in\mathbb{N}$
(or $\min(d)\in\mathbb{N}$ and $\max(d)\in\mathbb{N}$) denote 
the minimum and maximum number of lots (or processing time)
after which the maintenance operation has to be performed.
Finally, for any production operation $p[i]$, setup $s$, and maintenance 
operation $c$ or $d$,
$\mathrm{time}(p[i])\in\mathbb{N}$, $\mathrm{time}(s)\in\mathbb{N}$, $\mathrm{time}(c)\in\mathbb{N}$ or
$\mathrm{time}(d)\in\mathbb{N}$ provide the time required for performing
the respective operation or changing to the machine setup,
respectively.

The general properties above describe production routes and features of machines, and a set~$L$ of wafer lots represents the requested products,
where each lot $l\in L$ belongs to some product $p(l)\in P$.
A \emph{machine assignment} $m(l[1])\in M(p(l)[1]),\linebreak[1]\dots,\linebreak[1]m(l[n_{p(l)}])\in M(p(l)[n_{p(l)}])$
determines a specific machine to perform each operation $l[i]$ in the production route for a lot~$l$.
The \emph{schedule} for a machine $m$ in the tool group $M$ is a finite sequence
$m[1],\dots,m[n_m]$ of sets of operations,
where for each $1\leq j\leq n_m$:\pagebreak[1]
\begin{equation*}
m[j]=
\begin{cases}
\{l_1[i],\dots,l_k[i]\} &
\text{for lots } \{l_1,\dots,l_k\}\subseteq L
\text{ with } 
p(l_1) = \ldots = p(l_k) = p,
{} \\ &
i \leq n_p, m(l_1[i]) = \ldots = m(l_k[i]) = m, k\leq \max(p[i])
\text{;}
\\
\{s\} & 
\text{for some setup } s>0
\text{;} 
\\
\{c\} &
\text{for some maintenance operation } c\in c(M)
\text{; or}
\\
\{d\} &
\text{for some maintenance operation } d\in d(M)
\text{.}
\end{cases}
\end{equation*}
Starting from the initial machine setup $s(m)[1]=0$, we define the successor setups
for $1<j\leq n_m$ by $s(m)[j]=\{s\}$ if
$m[j-1]=\{s\}$ indicates a change to the setup $s\in\mathbb{N}$, or
$s(m)[j]=s(m)[j-1]$ otherwise.
Moreover, let $l(m[j])=\{l_1[i],\dots,l_k[i]\}$ if
$m[j]=\{l_1[i],\linebreak[1]\dots,\linebreak[1]l_k[i]\}$ for lots $\{l_1,\dots,l_k\}\subseteq L$ whose $i$-th operation is processed in batch,
or $l(m[j])=\emptyset$ otherwise.
The schedule for $m$ is \emph{feasible} if each $l[i]$ with $m(l[i])=m$ belongs to exactly one set $m[j]$ of operations, and for each $1\leq j\leq n_m$:
\begin{itemize}
\item $s(m)[j]=s(p(l)[i])$ if $s(p(l)[i])>0$ for some lot $l\in L$ with $l[i]\in l(m[j])$,
\item $\sum_{\max(\{0\}\cup\{{j_c}<j \mid m[{j_c}]=\{c\})< j' \leq j}|l(m[j'])| \leq \max(c)$
for each $c\in c(M)$, 
\item $\min(c) \leq \sum_{\max(\{0\}\cup\{{j_c}<j \mid m[{j_c}]=\{c\})< j' < j}|l(m[j'])|$
if $m[j]=\{c\}$ for $c\in\nolinebreak c(M)$,
\item $\sum_{\max(\{0\}\cup\{{j_d}<j \mid m[{j_d}]=\{d\})< j' \leq j,l[i]\in l(m[j'])}(\mathrm{time}(p(l)[i])\div|l(m[j'])|) \leq \max(d)$
for each $d\in d(M)$, and
\item $\min(d) \leq \sum_{\max(\{0\}\cup\{{j_d}<j \mid m[{j_d}]=\{d\})< j' < j,l[i]\in l(m[j'])}(\mathrm{time}(p(l)[i])\div|l(m[j'])|)$
if $m[j]=\{d\}$ for $d\in\nolinebreak d(M)$.
\end{itemize}
That is, the required (positive) setup must be in place when performing a production operation, and the number of lots (or processing time)
between maintenance operations $c\in c(M)$ (or $d\in d(M)$) must lie
in the range $[\min(c),\max(c)]$
(or $[\min(d),\max(d)]$).

Given a feasible schedule for each machine~$m$,
for each $1\leq j\leq n_m$,
we denote the \emph{operation time} of $m[j]$ by
$o(m[j])=\mathrm{time}(p(l)[i])$ if there is some $l[i]\in l(m[j])$, or
$o(m[j])=\mathrm{time}(o)$ if $m[j]\setminus l(m[j])=\{o\}$. 
Then, starting from $o(m[0])=0$ and $t(m[0])=\nolinebreak 0$,
the earliest \emph{start time} of $m[j]$ is
\begin{equation*}
t(m[j])=\max\left(
	         \begin{array}{@{}l@{}}
	         \{t(m[j{-}1])+o(m[j{-}1])\}\cup {} \\
             \{t(m'[j'])+o(m'[j']) \mid l[i]\in l(m[j]),1< i,l[i{-}1]\in l(m'[j'])\}
			 \end{array}
			\right)
\text{.}
\end{equation*}
The start time $t(m[j])$ thus reflects the earliest time at which
$m[j{-}1]$ is completed by machine~$m$ and
the predecessor operations $l[i{-}1]$ (if any) of all $l[i]\in l(m[j])$ 
have been finished as well.
Note that start times become infinite when the schedules for machines induce
circular waiting dependencies between the production operations for lots,
and we say that the (global) schedule of machine assignments for lots and
feasible schedules for machines is \emph{globally feasible} if all start times are finite.

The \emph{makespan} of a globally feasible schedule is the maximum
completion time $t(m[n_m])+o(m[n_m])$ over all machines~$m$.
An operation $m[j_s]=\{s\}$ constitutes a \emph{setup violation} for
$s\in\mathbb{N}$ if 
$m[j]\in\mathbb{N}$ for some $j>j_s$ indicates a setup change such that
$|\{j_s<j'<\nolinebreak j \mid l(m[j'])\neq\emptyset\}|<\min(s)$.
Moreover, $m[j]$ amounts to a \emph{batch violation} if we have that
$|l(m[j])| < \min(p(l)[i])$ for some $l[i]\in l(m[j])$.
The makespan, setup and batch violations provide
optimization objectives to be minimized for globally feasible schedules.

\begin{figure}[t]
	\centering
	\begin{tikzpicture}[thick, 
		x=0.75ex,y=3.5ex,label distance=-1ex
		]
		\SetScales
		
	    \foreach \i in {1,...,1} {
	    	\node[label=left:\textbf{implant\_128}] at (0,3.5-\i) {};
	    }
	    \foreach \i in {1,...,1} {
	    	\node[label=left:\textbf{lithotrack\_fe\_95}] at (0,3.5-\i-1) {};
	    }
	    \foreach \i in {1,...,1} {
	    	\node[label=left:\textbf{diffusion\_fe\_120}] at (0,3.5-\i-2) {};
	    }
	    \foreach \i in {0,10,20,30,40,50,60,70,80,90} {
	    	\draw[dotted] (\i,0) -- (\i,3) node [below] at (\i,0) {$\i$};
	    }
	    \foreach \i in {0,...,3} {
	    	\draw[dotted] (0,\i) -- (90,\i);
	    }
		
		\node[minimum width=20\scaledx-0.05\scaledx,minimum height=1\scaledy-0.05\scaledy,inner sep=0,rectangle,draw,anchor=south west,pattern color=lightgray,pattern=grid] at (0,0) {\textbf{[1,2],1}}; 
		
		\node[minimum width=22\scaledx-0.05\scaledx,minimum height=1\scaledy-0.05\scaledy,inner sep=0,rectangle,draw,anchor=south west,pattern color=lightgray,pattern=crosshatch] at (0,1) {\textbf{su450\_3}};
		
		\node[minimum width=6\scaledx-0.05\scaledx,minimum height=1\scaledy-0.05\scaledy,inner sep=0,rectangle,draw,anchor=south west,pattern color=lightgray,pattern=vertical lines] at (22,1) {\textbf{2,2}}; 
		
		\node[minimum width=6\scaledx-0.05\scaledx,minimum height=1\scaledy-0.05\scaledy,inner sep=0,rectangle,draw,anchor=south west,pattern color=lightgray,pattern=horizontal lines] at (28,1) {\textbf{1,2}}; 
		
		\node[minimum width=20\scaledx-0.05\scaledx,minimum height=1\scaledy-0.05\scaledy,inner sep=0,rectangle,draw,anchor=south west,pattern color=lightgray,pattern=crosshatch] at (8,2) {\textbf{su128\_1}};
		
		\node[minimum width=8\scaledx-0.05\scaledx,minimum height=1\scaledy-0.05\scaledy,inner sep=0,rectangle,draw,anchor=south west,pattern color=lightgray,pattern=vertical lines] at (28,2) {\textbf{2,3}}; 
		
		\node[minimum width=8\scaledx-0.05\scaledx,minimum height=1\scaledy-0.05\scaledy,inner sep=0,rectangle,draw,anchor=south west,pattern color=lightgray,pattern=horizontal lines] at (36,2) {\textbf{1,3}}; 
		
		\node[minimum width=15\scaledx-0.05\scaledx,minimum height=1\scaledy-0.05\scaledy,inner sep=0,rectangle,draw,anchor=south west,pattern color=lightgray,pattern=bricks] at (34,1) {\textbf{wk}};
		
		\node[minimum width=5\scaledx-0.05\scaledx,minimum height=1\scaledy-0.05\scaledy,inner sep=0,rectangle,draw,anchor=south west,pattern color=lightgray,pattern=horizontal lines] at (49,1) {\textbf{1,4}}; 
		
		\node[minimum width=5\scaledx-0.05\scaledx,minimum height=1\scaledy-0.05\scaledy,inner sep=0,rectangle,draw,anchor=south west,pattern color=lightgray,pattern=vertical lines] at (54,1) {\textbf{2,4}}; 
		
		\node[minimum width=18\scaledx-0.05\scaledx,minimum height=1\scaledy-0.05\scaledy,inner sep=0,rectangle,draw,anchor=south west,pattern color=lightgray,pattern=crosshatch] at (44,2) {\textbf{su128\_2}};
		
		\node[minimum width=7\scaledx-0.05\scaledx,minimum height=1\scaledy-0.05\scaledy,inner sep=0,rectangle,draw,anchor=south west,pattern color=lightgray,pattern=vertical lines] at (62,2) {\textbf{2,5}}; 
		
		\node[minimum width=13\scaledx-0.05\scaledx,minimum height=1\scaledy-0.05\scaledy,inner sep=0,rectangle,draw,anchor=south west,pattern color=lightgray,pattern=bricks] at (69,2) {\textbf{mn}};
		
		\node[minimum width=7\scaledx-0.05\scaledx,minimum height=1\scaledy-0.05\scaledy,inner sep=0,rectangle,draw,anchor=south west,pattern color=lightgray,pattern=horizontal lines] at (82,2) {\textbf{1,5}}; 
	
	\end{tikzpicture}
	\figspace\figspace
	\caption{
	         The chart illustrates an optimal schedule for an example SMSP instance with two lots of the same product, indicated by the labels $1$ and $2$ followed by respective production operation numbers from $1$ to $5$.
			 The production operations are performed by machines in three tool groups, called \emph{implant\_128}, \emph{lithotrack\_fe\_95}, and \emph{diffusion\_fe\_120},
			 with $1$ machine in each.
			 The \emph{diffusion\_fe\_120} machine starts by processing
			 the first operation for the batch of both lots,
			 while the remaining four operations per lot are performed
			 sequentially by the \emph{lithotrack\_fe\_95} and \emph{implant\_128} machines.
			 The \emph{su450\_3}, \emph{su128\_1}, and \emph{su128\_2} slots indicate the equipping of machines with required setups,
			 and the additional \emph{wk} and \emph{mn} slots denote maintenance operations.\label{fig:schedule}}
\end{figure}
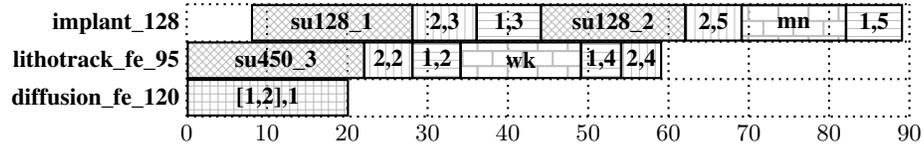
For example, an (optimal) schedule for an SMSP instance is displayed in Figure~\ref{fig:schedule}.
The machine in the \emph{diffusion\_fe\_120} tool group is capable of batching and processes the first operation in the route of two lots of the same product simultaneously.
Meanwhile, the setups \emph{su450\_3} and \emph{su128\_1}, required for sequential successor operations on machines in the tool groups \emph{lithotrack\_fe\_95} and \emph{implant\_128}, are brought in place before
processing the second and third production operations for each lot.
The machine in the \emph{lithotrack\_fe\_95} group undergoes a maintenance operation labeled \emph{wk} and then continues with the fourth operation in the production route for both lots.
The fifth and last operation per lot is processed by the machine in the tool group \emph{implant\_128}, where a switch to setup \emph{su128\_2} as well as a maintenance operation labeled \emph{mn} need to be performed in addition.
The makespan $89$ of this schedule is optimal, and likewise the setup and batch operations, the machine assignment of operations is fixed for simplicity, yet revisits of the \emph{lithotrack\_fe\_95} and \emph{implant\_128} machines illustrate re-entrant flow.

\section{Hybrid ASP encoding}\label{sec:asp}
In the following, we present our hybrid ASP with difference logic encoding
of SMSP supplying multi-objective optimization functionalities.
We start by describing the fact format of problem instances
(Section~\ref{subsec:instance}), followed by static preallocation strategies to
limit the number of assignable machines for each operation
(Section~\ref{subsec:partial}), then
the main encoding part to generate schedules incorporating batches, setup and maintenance operations (Section~\ref{subsec:schedule}), 
and finally optimization by multi-shot ASP solving
(Section~\ref{subsec:optimization}).

\subsection{Problem instance}\label{subsec:instance}

The example SMSP instance for which an (optimal) schedule is shown in
Figure~\ref{fig:schedule} is represented by the facts in Listing~\ref{prg:facts}. The structure of the used predicates is as follows:
\begin{description}[font=\normalfont\ttfamily\footnotesize]
\item[route($p$,$i$,$g$,$t$,$m$,$n$,$s$).]
The $i$-th operation for lots of product $p$ takes the processing time $t$ on a machine in tool group $g$ equipped with the setup $s$, where $m$ and $n$ provide the minimum and maximum batch size.
E.g., the fact in line~\ref{prg:facts:prd:begin} of Listing~\ref{prg:facts} states that the first operation for lots of product $1$ needs to be performed by a machine in the \emph{diffusion\_fe\_120} tool group, taking the processing time~$20$ for batches of (preferably) at least $2$ and at most $4$ lots with an arbitrary setup in view of $s=0$.
\item[setup($g$,$s$,$t$,$m$).]
Changing to setup $s\neq 0$ takes time $t$ on machines of the tool group $g$,
and at least $m$ operations should be processed in the setup $s$ before performing
another setup change.
The facts in lines~\ref{prg:facts:set:begin} and~\ref{prg:facts:set:mid} of Listing~\ref{prg:facts} express that the setups \emph{su128\_1} and \emph{su128\_2} need $20$ or $18$ time units, respectively, to be equipped on machines in the \emph{implant\_128} tool group,
where each of them ought to be maintained for $4$ production operations at minimum before changing.
Unlike that, the \emph{su450\_3} setup, taking $22$ units for equipping \emph{lithotrack\_fe\_95} machines with it, can be changed freely, as declared by the fact in line~\ref{prg:facts:set:end}.
\item[pm($g$,$l$,$e$,$m$,$n$,$t$).]
Machines in the tool group $g$ need to undergo a periodic maintenance operation labeled $l$,
whose type $e$ is either \lstinline{lots} or \lstinline{time}, the parameters $m$ and $n$
specify a minimum and maximum amount of lots or processing time, respectively, after which the
maintenance operation taking $t$ time units needs to repeated.
The facts in lines \ref{prg:facts:pm:begin}-\ref{prg:facts:pm:end} of Listing~\ref{prg:facts}
specify one maintenance operation per tool group, where \emph{implant\_128\_mn} is based on processed lots, while \emph{lithotrack\_fe\_95\_wk} and \emph{diffusion\_fe\_120\_mn}
need to repeated according to accumulated processing times.
\item[tool($g$,$l$).]
A machine labeled $l$ belongs to the tool group~$g$,
where the facts in line~\ref{prg:facts:tool:begin} of Listing~\ref{prg:facts}
introduce one machine per tool group.
\item[lot($l$,$p$).]
A lot labeled~$l$ of product~$p$ needs to produced,
and two lots of the (single) product~\lstinline{1} are
declared by the facts in line~\ref{prg:facts:wip:begin} of Listing~\ref{prg:facts}.
\end{description}%
\begin{lstlisting}[float=t,label=prg:facts,caption={Facts for an SMSP instance with two lots and three tool groups with one machine each},linerange={2-14}]
route(1, 1, diffusion_fe_120, 20, 2, 4,       0). #(\label{prg:facts:prd:begin}#)
route(1, 2, lithotrack_fe_95,  6, 1, 1, su450_3).
route(1, 3,      implant_128,  8, 1, 1, su128_1).
route(1, 4, lithotrack_fe_95,  5, 1, 1, su450_3).
route(1, 5,      implant_128,  7, 1, 1, su128_2). #(\label{prg:facts:prd:end}#) 
setup(     implant_128, su128_1, 20, 4). #(\label{prg:facts:set:begin}#)
setup(     implant_128, su128_2, 18, 4). #(\label{prg:facts:set:mid}#)
setup(lithotrack_fe_95, su450_3, 22, 0). #(\label{prg:facts:set:end}#)
pm(     implant_128,      implant_128_mn, lots,   2,   3, 13). #(\label{prg:facts:pm:begin}#)
pm(lithotrack_fe_95, lithotrack_fe_95_wk, time,  12,  20, 15). 
pm(diffusion_fe_120, diffusion_fe_120_mn, time, 125, 150, 35). #(\label{prg:facts:pm:end}#)
tool(implant_128, 1). tool(lithotrack_fe_95, 1). tool(diffusion_fe_120, 1). #(\label{prg:facts:tool:begin}#) #(\label{prg:facts:tool:end}#)
lot(1, 1). lot(2, 1). #(\label{prg:facts:wip:begin}#) #(\label{prg:facts:wip:end}#)
\end{lstlisting}

\subsection{Partially flexible machine assignment}\label{subsec:partial}

\begin{lstlisting}[float=t,label=prg:groups,caption={Encoding part for partitioning tool groups into subgroups and preallocation by setups},linerange={1-45,65-69}]
#const sub_size = 0.  % if "sub_size" is positive, it limits machines in subgroups #(\label{prg:encoding:1}#)
#const lot_step = 0.  % if "lot_step" is set to 1, operations of lot get separated #(\label{prg:encoding:2}#)
#const by_setup = 0.  % if "by_setup" is set to 1, operation setups get subdivided #(\label{prg:encoding:3}#)

% split tool groups into subgroups of at most "sub_size" many machines in each

subgroup_size(sub_size) :- sub_size > 0. #(\label{prg:encoding:7}#)

subgroup(G, 1, M)      :- tool(G, M), sub_size = 0. #(\label{prg:encoding:9}#)
subgroup(G, J, M)      :- tool(G, N), subgroup_size(S), not tool(G, N+1), #(\label{prg:encoding:10}#)
                          J = 1 .. K, K = (N+S-1)/S, L = (N\K+J-|N\K-J|)/2,
                          M = (J-1)*(N/K)+(L+J-1-|L-J+1|)/2+1 .. J*(N/K)+L. #(\label{prg:encoding:12}#)
subgroup(G, J, M1, M2) :- subgroup(G, J, M1), not subgroup(G, J, M1-1), #(\label{prg:encoding:13}#)
                          subgroup(G, J, M2), not subgroup(G, J, M2+1). #(\label{prg:encoding:14}#)

% map each production operation to some subgroup of the respective tool group

step_index(G, K, (L, P, I, T, S)) :- lot(L, P), route(P, I, G, T, M, N, S), #(\label{prg:encoding:18}#)
                             K = #count{P', L', I' * lot_step : lot(L', P'),
                                           route(P', I', G, T', M', N', S'),
                             (P', L', I' * lot_step) < (P, L, I * lot_step)}. #(\label{prg:encoding:21}#)

step_subgroup(G, K\J+1, O) :- step_index(G, K, O), #(\label{prg:encoding:23}#)
                              subgroup(G, J, _), not subgroup(G, J+1, _). #(\label{prg:encoding:24}#)

assignable(O, G, M) :- step_subgroup(G, J, O), subgroup(G, J, M), by_setup = 0. #(\label{prg:encoding:26}#)

% distinguish machines in each subgroup based on production operations' setups

subgroup_setup(G, J, S)    :- step_subgroup(G, J, (L, P, I, T, S)), by_setup != 0. #(\label{prg:encoding:30}#)
subgroup_setup(G, J, S, N) :- subgroup_setup(G, J, S), #(\label{prg:encoding:31}#)
                     N = #sum+{T, L, P, I : step_subgroup(G, J, (L, P, I, T, S))}. #(\label{prg:encoding:32}#)

setup_index(G, J, K, S, N) :- subgroup_setup(G, J, S, N), #(\label{prg:encoding:34}#)
          K = #count{S', N' : subgroup_setup(G, J, S', N'), (N, S) < (N', S')}. #(\label{prg:encoding:35}#)
setup_index(G, J, K, N)    :- setup_index(G, J, K, S, N). #(\label{prg:encoding:36}#)
setup_index(G, J, K)       :- setup_index(G, J, K-1, _), #(\label{prg:encoding:37}#)
                              not setup_index(G, J, K, _). #(\label{prg:encoding:38}#)

setup_machine(G, J, M, M\K, N) :- subgroup(G, J, M1, M2), setup_index(G, J, K), #(\label{prg:encoding:40}#)
                                  setup_index(G, J, M\K, N), M = 0 .. M2-M1. #(\label{prg:encoding:41}#)
setup_machine(G, J, M, K, A+N) :- subgroup(G, J, M1, M2), setup_index(G, J, K, N), #(\label{prg:encoding:42}#)
                                  next_machine(G, J, M2-M1, K-1, A, M). #(\label{prg:encoding:43}#)

%* [...] *% #(\label{prg:encoding:45}#)

assignable((L, P, I, T, S), G, M1+M) :- step_subgroup(G, J, (L, P, I, T, S)), #(\label{prg:encoding:47}#)
                                        setup_machine(G, J, M, K, A),
                                        setup_index(G, J, K, S, N),
                                        subgroup(G, J, M1, M2). #(\label{prg:encoding:50}#)
\end{lstlisting}
Experiments with our prototypical SMSP encoding \cite{ali2023flexible} showed that fixing the machine assignment of operations upfront sacrifices optimality, while a fully flexible assignment leads to plenty ground rules slowing down the optimization when a tool group contains many machines.
To enable trade-offs between the fixed and fully flexible machine allocation strategies, the novel encoding part in Listing~\ref{prg:groups} introduces a constant
\lstinline{sub_size} that allows for limiting the number of assignable machines per operation.
That is, when \lstinline{sub_size} is \lstinline{0},
the machine assignment remains fully flexible, gets fixed if the value is~\lstinline{1},
or is limited to some \emph{subgroup} of a tool group with at most \lstinline{sub_size} many machines for values greater than one.
In the latter case, the rule in lines \ref{prg:encoding:10}-\ref{prg:encoding:12}
partitions \lstinline{N} machines of a tool group \lstinline{G} into
$\lceil\text{\lstinline{N}}\div\text{\lstinline{sub_size}}\rceil$ many subgroups,
each gathering \lstinline{sub_size} or $\text{\lstinline{sub_size}}-1$ of the machines in \lstinline{G} when $\text{\lstinline{N}}\geq \text{\lstinline{sub_size}}$.
For example, we derive the atoms
\lstinline{subgroup(implant_128,1,1,3)},
\lstinline{subgroup(implant_128,2,4,5)}, and
\lstinline{subgroup(implant_128,}\linebreak[1]\lstinline{3,}\linebreak[1]\lstinline{6,7)}
by the rule in lines~\ref{prg:encoding:13}-\ref{prg:encoding:14},
giving the subgroups
$\{\text{\lstinline{1}},\text{\lstinline{2}},\text{\lstinline{3}}\}$,
$\{\text{\lstinline{4}},\text{\lstinline{5}}\}$, and
$\{\text{\lstinline{6}},\text{\lstinline{7}}\}$
when
seven machines in the \emph{implant\_128} tool group are partitioned for
the \lstinline{sub_size} value~\lstinline{3}.

The rules in lines \ref{prg:encoding:18}-\ref{prg:encoding:24} determine the subgroup
to which an operation is allocated, based on a lexicographical index
for operations to be processed by machines in the same tool group.
This allocation can be configured by the constant \lstinline{lot_step}: if its value is \lstinline{0}, all operations of a lot are mapped to a common index, or to successive indexes in case of value~\lstinline{1}.
The rationale for these two strategies is that operations performed on the same lot
succeed one another and will thus never compete for a machine.
On the other hand, the operations may require different setups so that changes are needed when
reusing the same machine.
In fact, the latter indexing strategy is likely to map operations of a lot to separate subgroups,
as the rule in lines \ref{prg:encoding:23}-\ref{prg:encoding:24} allocates them in a round robin fashion.

As subordinate machine allocation criterion within each subgroup, the setups of operations can be inspected by means of the rules in lines \ref{prg:encoding:30}-\ref{prg:encoding:50} when the constant \lstinline{by_setup} is set to a value other than~\lstinline{0}.
The idea of the rules in lines \ref{prg:encoding:30}-\ref{prg:encoding:38}
is to order setups by the sum of processing times for their operations, where setups requiring more processing time come first.
Then the rules in lines \ref{prg:encoding:40}-\ref{prg:encoding:50} follow this order to map setups and the respective operations to specific machines, always picking the machine with the least load so far for the next setup to allocate.
The rules to determine the least loaded machine for a setup are omitted in Listing~\ref{prg:groups} to save space, and our full encoding is available online.%
\footnote{\url{https://github.com/prosysscience/FJSP-SMT2020}\label{foo:online}}
For example, if two machines of the \emph{implant\_128} tool group belong to the same subgroup for
the lots specified by the facts in Listing~\ref{prg:facts}, the allocation by setups yields the atoms
\lstinline{assignable((1,1,3,8,}\linebreak[1]\lstinline{su128_1),}\linebreak[1]\lstinline{implant_128,}\linebreak[1]\lstinline{1)},
\lstinline{assignable((2,1,3,8,}\linebreak[1]\lstinline{su128_1),}\linebreak[1]\lstinline{implant_128,}\linebreak[1]\lstinline{1)},
\lstinline{assignable((1,1,5,7,}\linebreak[1]\lstinline{su128_2),}\linebreak[1]\lstinline{implant_128,}\linebreak[1]\lstinline{2)}, and
\lstinline{assignable((2,1,5,7,}\linebreak[1]\lstinline{su128_2),}\linebreak[1]\lstinline{implant_128,}\linebreak[1]\lstinline{2)},
thus mapping the third operation in the route of both lots, requiring the setup \emph{su128\_1}, to the first 
and the fifth operation for both with the setup \emph{su128\_2} to the second machine of the subgroup.

\subsection{Schedule generation}\label{subsec:schedule}

While the previous encoding part specifies preallocation strategies to
statically limit the machines to which each operation may be assigned,
Listing~\ref{prg:assign} describes the actual, combinatorial scheduling task,
including the machine assignment, setup and maintenance operations, as well as
the aggregation of \emph{batches}.
The latter feature was not yet incorporated in our prototypical SMSP encoding
\cite{ali2023flexible} and is newly introduced by the rules in lines
\ref{prg:encoding:54}-\ref{prg:encoding:64}.
To begin with, (ordered) pairs of operations processed by machines with a
maximum batch size beyond one are determined in lines
\ref{prg:encoding:54}-\ref{prg:encoding:58}.
Then, batches are generated by applying the choice rule in line~\ref{prg:encoding:60}, which represents batch processing of the first operation
for both lots given by the facts in Listing~\ref{prg:facts} in terms of the derivable atom
\lstinline{batch(diffusion_fe_120,(1,1,1,20,0),(2,1,1,20,0))}.
That is, a batch is identified by the operation on its lexicographically smallest lot, to which lots with greater identifiers are linked via the \lstinline{batch/3} predicate.
Any such linked lots are indicated by \lstinline{batched/1},
and \lstinline{batched/2} provides a symmetric version of \lstinline{batch/3},
where both of the \lstinline{batched} predicates are derived by the rule in
line~\ref{prg:encoding:63}.
The integrity constraint in line~\ref{prg:encoding:64} makes sure that batches
partition the lots of a product, as it rules out that the lot identifying a batch is itself linked to another (lexicographically smaller) lot.
This unambiguous batch representation is exploited by the integrity constraint in line~\ref{prg:encoding:61}, where it suffices to count the linked lots to assert that the maximum batch size for an operation is not exceeded.
\begin{lstlisting}[float=t,label=prg:assign,caption={Encoding part to assign batches, machines, as well as setup and maintenance operations},linerange={71-109,120-129,150-156},firstnumber=52]
% generate batches for operations of different lots with same product in subgroups

step_batch(G, J, M, N, (L, P, I, T, S)) :- #(\label{prg:encoding:54}#)
      step_subgroup(G, J, (L, P, I, T, S)), route(P, I, G, T, M, N, S), 1 < N. #(\label{prg:encoding:55}#)
lots_batch((L1, P, I, T, S), (L2, P, I, T, S)) :- #(\label{prg:encoding:56}#)
      step_batch(G, J, M, N, (L1, P, I, T, S)),
      step_batch(G, J, M, N, (L2, P, I, T, S)), L1 < L2. #(\label{prg:encoding:58}#)

{batch(G, O1, O2) : lots_batch(O1, O2)} 1 :- step_batch(G, J, M, N, O2). #(\label{prg:encoding:60}#)
:- step_batch(G, J, M, N, O1), N #count{O2 : batch(G, O1, O2)}. #(\label{prg:encoding:61}#)

batched(O1, O2; O2, O1; O2) :- batch(G, O1, O2). #(\label{prg:encoding:63}#)
:- batched(O1), batch(G, O1, O2). #(\label{prg:encoding:64}#)

% generate assignment of production operations to machines and order of processing

{assign(O, G, M) : assignable(O, G, M)} = 1 :- step_index(G, K, O). #(\label{prg:encoding:68}#)

step_pair(G, O1, O2)   :- assignable(O1, G, M), assignable(O2, G, M), O1 < O2. #(\label{prg:encoding:70}#)
step_assign(G, O1, O2) :- step_pair(G, O1, O2), assign(O1, G, M), assign(O2, G, M). #(\label{prg:encoding:71}#)
lots_assign(G, O1, O2) :- O1 = (L1, P1, I1, T1, S1), step_assign(G, O1, O2), #(\label{prg:encoding:72}#)
                          O2 = (L2, P2, I2, T2, S2), (L1, P1) < (L2, P2). #(\label{prg:encoding:73}#)
in_sequence(G, O1, O2) :- lots_assign(G, O1, O2), not batch(G, O1, O2). #(\label{prg:encoding:74}#)
:- batch(G, O1, O2), not lots_assign(G, O1, O2). #(\label{prg:encoding:75}#)

{lots_order(G, O1, O2)} :- in_sequence(G, O1, O2). #(\label{prg:encoding:77}#)
 lots_order(G, O2, O1)  :- in_sequence(G, O1, O2), not lots_order(G, O1, O2). #(\label{prg:encoding:78}#)

% determine setup and maintenance operations required before production operations

main_setup(G, H, K, W) :- pm(G, H, E, M, N, W), #(\label{prg:encoding:82}#)
          K = #count{H' : pm(G, H', E', M', N', W'), (W, H) <= (W', H')}. #(\label{prg:encoding:83}#)
main_setup(G, S, 0, U) :- setup(G, S, U, M). #(\label{prg:encoding:84}#)

step_order(G, O1, O2) :- main_setup(G, X, K, V), lots_order(G, O1, O2). #(\label{prg:encoding:86}#)
step_order(G, O1, O2) :- main_setup(G, X, K, V), step_assign(G, O1, O2), #(\label{prg:encoding:87}#)
                         O1 = (L, P, I1, T1, S1), O2 = (L, P, I2, T2, S2). #(\label{prg:encoding:88}#)

%* [...] *% #(\label{prg:encoding:90}#)

equip(O, G, S) :- setup(G, S, U, M), step_index(G, K, O), O = (L, P, I, T, S), #(\label{prg:encoding:92}#)
                  not reuse(G, O). #(\label{prg:encoding:93}#)

{maintain(O, G, H)} :- pm(G, H, E, M, N, W), step_index(G, K, O). #(\label{prg:encoding:95}#)
:- batched(O1, O2), maintain(O1, G, H), not maintain(O2, G, H). #(\label{prg:encoding:96}#)
:- pm(G, H, E, M, N, W), step_index(G, K, O), not step_order(G, _, O), #(\label{prg:encoding:97}#)
   not maintain(O, G, H). #(\label{prg:encoding:98}#)

%* [...] *% #(\label{prg:encoding:100}#)

delay(O, G, 0, U)   :- main_setup(G, S, 0, U), equip(O, G, S). #(\label{prg:encoding:116}#)
delay(O, G, 0, U+V) :- main_setup(G, S, 0, U), equip(O, G, S), delay(O, G, 1, V). #(\label{prg:encoding:117}#)
delay(O, G, K, W)   :- main_setup(G, H, K, W), maintain(O, G, H), 0 < K. #(\label{prg:encoding:118}#)
delay(O, G, K, W+V) :- main_setup(G, H, K, W), maintain(O, G, H), 0 < K, #(\label{prg:encoding:119}#)
                       delay(O, G, K+1, V). #(\label{prg:encoding:120}#)
delay(O, G, K-1, V) :- delay(O, G, K, V), 0 < K.
\end{lstlisting}

The choice rule in line~\ref{prg:encoding:68} continues with the \emph{machine assignment} by selecting exactly one machine, among those determined by a preallocation strategy from the previous subsection, for processing an operation.
Operation pairs assigned to the same machine are brought into an ordered
representation in terms of the \lstinline{step_assign/3} predicate via the rules in lines
\ref{prg:encoding:70}-\ref{prg:encoding:71}.
These pairs are filtered in lines \ref{prg:encoding:72}-\ref{prg:encoding:73} to
indicate the operations on different lots by \lstinline{lots_assign/3}.
Only for the latter an execution order needs to be guessed by applying the rules
in lines \ref{prg:encoding:77}-\ref{prg:encoding:78},
provided that the operations do not belong to the same batch,
which is checked in line~\ref{prg:encoding:74}.
However, the operations in a batch must share a common machine, as asserted by the integrity constraint in line~\ref{prg:encoding:75}.

The execution order of operations sharing a machine must be inspected further to
allocate required \emph{setup and maintenance} operations.
As several kinds of periodic maintenance may need to be applied to machines of the same tool group and their durations add up when they are performed in sequence,
the rules in lines \ref{prg:encoding:82}-\ref{prg:encoding:84} associate
maintenance operations with (positive) indexes in decreasing order of their
durations, with the additional index \lstinline{0} used for operation setups other
than (don't care) setup~$0$.
E.g., the facts in Listing~\ref{prg:facts} yield the atoms
\lstinline{main_setup(implant_128,implant_128_mn,1,13)},
\lstinline{main_setup(implant_128,su128_1,0,20)}, and
\lstinline{main_setup(implant_128,su128_2,0,18)} in view of the
periodic \emph{mn} maintenance along with the \emph{su128\_1} and
\emph{su128\_2} setups of operations processed by machines in the
\emph{implant\_128} tool group.
For tracking the exact execution order of operations on a machine, also if they
involve the same lot,
the \lstinline{step_order/3} predicate determined by 
the rules in lines \ref{prg:encoding:86}-\ref{prg:encoding:88} augments the
guessed predicate \lstinline{lots_order/3} with atoms reflecting the production route of a lot revisiting the same machine.
While we omit the details to save space,
let us mention that the necessity of a setup change before performing an operation
is a consequence of the execution sequence on a machine, i.e., the rule in lines
\ref{prg:encoding:92}-\ref{prg:encoding:93} derives an atom of the
\lstinline{equip/3} predicate whenever the setup required for an operation is not
already in place.
Unlike that, maintenance procedures are subject to a range, either in terms of
processed lots or accumulated processing time, after which they have to be repeated.
Hence, the rule in line~\ref{prg:encoding:95} introduces the choice to perform a
specific maintenance before the next production operation, the integrity constraint in line~\ref{prg:encoding:96} distributes such a choice over all lots in a batch, and (auxiliary) maintenances before the first production operation on a machine are asserted in lines \ref{prg:encoding:97}-\ref{prg:encoding:98}.
The resulting maintenance and setup times needed before the next
production operation can be processed are then added up by the rules in
lines \ref{prg:encoding:116}-107, where additional rules and constraints ensuring the compliance of maintenance procedures to the specified repetition ranges are part of our full encoding.%
\footref{foo:online}
For example, the \emph{mn} maintenance performed before the fifth operation for 
lot~$1$ in Figure~\ref{fig:schedule} 
is expressed by the atoms
\lstinline{delay((1,1,5,7,su128_2),implant_128,1,13)} and
\lstinline{delay((1,1,5,7,su128_2),implant_128,0,13)},
the latter providing \lstinline{13} as the sum of
all maintenance and setup times required before the production operation can be processed.

\subsection{Multi-objective optimization}\label{subsec:optimization}

Our multi-objective optimization approach combines minimization at the level of 
difference logic variable values, as already used in \cite{ali2023flexible,el2022problem},
with native ASP optimization capacities, as applied in \cite{abels2021train,eiter2022answer,francescutto2021solving} w.r.t.\ the satisfaction of difference logic constraints,
by means of multi-shot solving functionalities~\cite{gekakasc17a}.
To this end, the rules in lines \ref{prg:encoding:125}-\ref{prg:encoding:138} of Listing~\ref{prg:optimize} assert difference logic constraints on the completion times of operations,
beginning with processing times of the first operations in production routes
(lines \ref{prg:encoding:125}-\ref{prg:encoding:126}) or processing plus setup times 
(lines \ref{prg:encoding:127}-\ref{prg:encoding:128}) for all operations to which the latter apply.
These lower bounds are propagated along the production route of each lot
(lines \ref{prg:encoding:131}-\ref{prg:encoding:132}) and
the processing order of operations 
on machines
(lines \ref{prg:encoding:133}),
where the times required for maintenance and setup are incorporated in addition
(lines \ref{prg:encoding:134}-\ref{prg:encoding:135}).
Notably, batches are handled by synchronizing the completion time between the
operations on involved lots in line~\ref{prg:encoding:130},
so that the predecessor operation (if any) finishing latest among all lots in the batch is decisive for
the entire batch.
The rule in lines
\ref{prg:encoding:137}-\ref{prg:encoding:138}
asserts the completion time of the last operation in each lot's production route as a
lower bound on the difference logic variable \lstinline{makespan}, thus enabling plain
\emph{makespan minimization} by supplying \lstinline{--minimize-variable=makespan} as an option to \clingodl.
\begin{lstlisting}[float=t,label=prg:optimize,caption={Encoding part for determining lot completion times and multi-objective optimization},linerange={158-191},firstnumber=109]
% assert difference logic constraints on completion times of production operations

&diff{0 - (L, P, 1, T, S)} <= -T   :- step_index(G, K, (L, P, 1, T, S)), #(\label{prg:encoding:125}#)
                                      not setup(G, S, _, _). #(\label{prg:encoding:126}#)
&diff{0 - (L, P, I, T, S)} <= -T-U :- step_index(G, K, (L, P, I, T, S)), #(\label{prg:encoding:127}#)
                                      setup(G, S, U, M). #(\label{prg:encoding:128}#)

&diff{O1 - O2} <= 0    :- batched(O1,O2). #(\label{prg:encoding:130}#)
&diff{O1 - O2} <= -T   :- step_index(G1, K1, O1), O1 = (L, P, I, T1, S1), #(\label{prg:encoding:131}#)
                          step_index(G2, K2, O2), O2 = (L, P, I+1, T, S). #(\label{prg:encoding:132}#)
&diff{O1 - O2} <= -T   :- lots_order(G, O1, O2), O2 = (L, P, I, T, S). #(\label{prg:encoding:133}#)
&diff{O1 - O2} <= -T-V :- step_order(G, O1, O2), O2 = (L, P, I, T, S), #(\label{prg:encoding:134}#)
                          delay(O2, G, 0, V). #(\label{prg:encoding:135}#)

&diff{O - makespan} <= 0 :- step_index(G, K, O), #(\label{prg:encoding:137}#)
                            O = (L, P, I, T, S), not route(P, I+1, _, _, _, _, _). #(\label{prg:encoding:138}#)

% subprograms for minimizing the makespan, setup and batch violations of schedules

#program opt(b). #(\label{prg:encoding:142}#)
#external bound(b). #(\label{prg:encoding:143}#)

&diff{makespan - 0} <= b :- bound(b). #(\label{prg:encoding:145}#)

#program weak(b). #(\label{prg:encoding:147}#)
bound(b). #(\label{prg:encoding:148}#)

change(G, O) :- change(G, O, O1). #(\label{prg:encoding:150}#)

:~ setup(G, S, U, M), equip(O, G, S), change(G, O), #(\label{prg:encoding:152}#)
   #count{O2 : reuse(G, O, O2), not batched(O2)} < M-1. [1@2, O] #(\label{prg:encoding:153}#)

:~ step_batch(G, J, M, N, O), not batched(O), #(\label{prg:encoding:155}#)
   #count{O2 : batch(G, O, O2)} M-2. [1@1, O] #(\label{prg:encoding:156}#)
\end{lstlisting}

However, to incorporate the minimization of \emph{setup and batch violations} as additional optimization criteria beyond makespan,
we utilize a custom control script on top of the Python interface of \clingodl.
Its first stage concerns makespan minimization,
where the \lstinline{opt(b)} subprogram in lines
\ref{prg:encoding:142}-\ref{prg:encoding:145} of Listing~\ref{prg:optimize}
is instantiated with the value $t-1$ for the parameter \lstinline{b} and then solved with the external atom \lstinline{bound(}$t-1$\lstinline{)} set to true
whenever an answer set such that \lstinline{makespan}${}=t$ has been found.
This makes sure that each answer set provides a schedule with strictly shorter
makespan until an unsatisfiable solving attempt yields that the makespan~$t$ of the last schedule is optimal.
In the latter case, the subprogram \lstinline{weak(b)} in lines
\ref{prg:encoding:147}-\ref{prg:encoding:156} gets instantiated with the 
value $t$ for \lstinline{b} (and possibly also \lstinline{opt(b)} if the value~$t$ has not been supplied for \lstinline{b} before),
which fixes the makespan of any subsequently found schedule to the optimum~$t$.
With the weak constraints in lines
\ref{prg:encoding:152}-\ref{prg:encoding:156} as well as the rule in
line~\ref{prg:encoding:150} at hand for indicating operations whose setup is
reinstalled after some temporary change,
the second stage consists of native ASP optimization for minimizing setup and batch violations.
Here we take setup violations, where a setup gets changed before performing the intended minimum number of production operations using it, as strictly more
significant (optimization level \lstinline{@2}) than violations of the minimum
batch size (optimization level \lstinline{@1}), considering that equipping a machine with a setup takes extra time and effort.
For example, the schedule in Figure~\ref{fig:schedule} involves one
setup violation due to changing from the setup \emph{su128\_1} to \emph{su128\_2}
before performing the intended minimum number of four operations with this setup
on the \emph{implant\_128} machine.
Since avoiding the setup violation would require a second machine in the \emph{implant\_128} tool group, the schedule is nevertheless optimal.

\section{Experiments}\label{sec:results}
\begin{table}[t]
	\centering
	\caption{Preallocation strategy results with $3$ machines per tool group and $10$ operations per lot}
	\label{tab:table}
	\figspace\scriptsize
		\begin{tabular}{|l
				cl||rr|rr|rr|rr|}
			\hline
			& \multicolumn{1}{@{\hspace{-3mm}}c@{\hspace{-3mm}}}{\textbf{9 Machines}}                   &                      & 
			\multicolumn{2}{r|}{\textbf{70 Operations}}                 & \multicolumn{2}{r|}{\textbf{80 Operations}}                 & \multicolumn{2}{r|}{\textbf{90 Operations}}                 & \multicolumn{2}{r|}{\textbf{100 Operations}}                 \\
			& Size 
			&        &
			Lot                         & Step                        & Lot                         & Step                        & Lot          & Step         & Lot          & Step         \\
			\hline\hline
			\multirow{3}{*}{\textbf{Fixed}}    & \multirow{3}{*}{1} & 
			Makespan    & 483                         & 428                         & 489                         & 440                         & 486          & 531          & 592          & \textbf{553}         \\
			&                    & 
			Setup/Batch & 6/12                        & 2/12                        & 5/14                        & 0/13                        & 5/14         & 3/12         & 3/12         & 0/16         \\
			&                    & 
			1\ts{st}/2\ts{nd} Stage & 2/1                         & TO/27                          & 6/2                        & TO/13                          & 11/13         & TO           & TO/78           & TO           \\
			\midrule
			\multirow{6}{*}{\textbf{Flexible}} & \multirow{3}{*}{2} & 
			Makespan    & 483                         & 475                         & 592                         & 592                         & 592          & 539          & 745          & 698          \\
			&                    & 
			Setup/Batch & 2/8                        & 0/9                        & 1/8                        & 1/8                        & 1/10         & 0/11          & 0/12          & 0/15          \\
			&                    & 
			1\ts{st}/2\ts{nd} Stage & 5/1                         & TO                          & TO/114                          & TO/1                          & TO/130           & TO           & TO           & TO          \\
			\cline{2-11}
			& \multirow{3}{*}{3} & 
			Makespan    & 559                         & --                          & 815                         & --                          & 1357 & -- & 1486 & -- \\ 
			&                    & 
			Setup/Batch & 0/8                         & --                          & 0/8                        & --                          & 0/10 & -- & 10/18 & -- \\ 
			&                    & 
			1\ts{st}/2\ts{nd} Stage & TO                       & --                          & TO/140                          & --                          & TO/79 & -- & TO & -- \\ 
			\midrule
			\multirow{6}{*}{\textbf{Setup}}    & \multirow{3}{*}{2} & 
			Makespan    & 483                         & 475                         & 592                         & 592                         & 592          & 536          & 745          & 683          \\
			&                    & 
			Setup/Batch & 2/8                        & 0/9                        & 1/8                        & 1/8                        & 1/10         & 0/12          & 0/13          & 0/16          \\
			&                    & 
			1\ts{st}/2\ts{nd} Stage & 2/1                        & TO                          & TO/21                          & TO/25                          & TO/22           & TO           & TO/76           & TO           \\
			\cline{2-11}
			& \multirow{3}{*}{3} & 
			Makespan    & \textbf{334}                         & --                          & \textbf{345}                         & --                          & \textbf{434}          & --           & 555          & --           \\
			&                    & 
			Setup/Batch & 0/8                         & --                          & 0/8                         & --                          & 0/11          & --           & 0/12          & --           \\
			&                    & 
			1\ts{st}/2\ts{nd} Stage & TO/20                       & --                          & TO/123                          & --                          & TO           & --           & TO/73           & --           \\
			\hline
		\end{tabular}
\end{table}
We constructed a scalable set of benchmark instances, focusing on sub-routes of
$10$ production operations for two product types from the SMT2020 simulation scenario~\cite{kopp2020smt2020}.
The $10$ operations in both sub-routes are processed by machines
belonging to three tool groups and do thus involve re-entrant flow,
as a lot visits the same tool group multiple times.
Moreover, the operations incorporate batching and specific setups, and machines undergo periodic maintenance operations.
In the following, we concentrate on instances with $9$ machines, i.e., $3$ per
tool group, and gradually increasing number of lots.
Further smaller- and larger-scale instances along with our implementation are
available online.\footref{foo:online}

We ran our experiments with \clingodl\ (version 1.4.0) on an Intel® Core™i7-8650U CPU Dell Latitude 5590 machine under Windows 10, imposing two time limits per run:
the first stage for makespan minimization is aborted at $450$ seconds, in which case the best schedule found so far 
is taken as upper bound on the makespan for proceeding to minimize setup and batch violations with 
another $150$ seconds time limit.

Table~\ref{tab:table} reports the quality of best schedules obtained within the time limits for both optimization stages, split into `Makespan' and `Setup/Batch'
values, while two runtimes or `TO' for a timeout, respectively, are given in the
`1\ts{st}/2\ts{nd} Stage' rows, only listing a single `TO' entry in case both stages timed out.
The `Size' column provides the value taken for the constant \lstinline{sub_size},
limiting the number of machines in subgroups to which the operations are preallocated.
For the latter, the `Lot' columns include results with value \lstinline{0} for the constant \lstinline{lot_step}, where a common subgroup takes all operations for a lot, or for value \lstinline{1} in the `Step' columns, leading to their distribution among subgroups.

The `Size' value 1 necessarily leads to a fixed machine assignment, for which the
quality indicators clearly show that the `Step' strategy yields better schedules,
although it incurs more timeouts and thus fewer certain optima because operations on different lots increase the flexibility of execution sequences and thus search complexity.
While flexibility within subgroups by setting their `Size' to 2 or 3 in principle allows for improved schedules, we observe a deterioration due to sharply increasing instantiation size and search effort, as already observed in \cite{ali2023flexible}.
The setup strategy to differentiate operations and machines within subgroups,
activated by changing the constant \lstinline{by_setup},
aims to cut down the scheduling complexity in line with the optimization objectives by reducing the need for setup changes.
This leads to significantly improved schedules with `Size' 3, where the
`Lot' and `Step' preallocation strategies are indifferent and redundant results for the latter are omitted, up to a critical size reached with $100$~operations.

With our preliminary approach~\cite{ali2023flexible}, using a more naive and less feature-rich encoding of either fixed or fully flexible machine assignments, the
threshold at which problem size and combinatorics get prohibitive was reached at less than $50$ operations already.
Despite gearing up to double that size, our benchmark instances still represent small excerpts of the large-scale semiconductor fabs with more than $100$ tool groups and from $242$ to $543$ production operations per lot modeled by~\cite{kopp2020smt2020}.
The elevated complexity in comparison to basic settings like the traditional FJSP is mainly due to sophisticated setup and maintenance operations, requiring a detailed analysis of execution sequences on machines for SMSP.
We conjecture that similar scalability limits would also be encountered with MIP or CP encodings, yet the first-order modeling language of ASP with difference logic facilitates rapid prototyping and experimentation.
In fact, our performance evaluation aims to explore the feasibility of search and optimization, in order to come up with strategies for breaking down large SMSP instances into more manageable portions, e.g., focusing on some bottleneck tool groups or re-entrant flow of operations.

\section{Conclusion}\label{sec:conclusion}
This work extends our preliminary SMSP approach~\cite{ali2023flexible}
with crucial features, namely, scalable and informed preallocation strategies to reduce the instantiation size and search complexity, as well as batch
processing and multiple optimization objectives.
While we enhance the scheduling scalability and coverage of real-world features,
our mid-term goal is to incorporate scheduling into the real or simulated management of semiconductor manufacturing processes.
As next step into this direction, we aim to use scheduling for improving the decision making in the PySCFabSim simulator~\cite{kotaalelgese22a}, where
methods available so far, i.e., handcrafted dispatching rules or black-box machine
learning models, function locally and do not take the global impact of their decisions into account.


\paragraph{Acknowledgments}
This work was funded by 
FFG project 894072 (SwarmIn) as well as
KWF project 28472, cms electronics GmbH, FunderMax GmbH, Hirsch Armbänder GmbH, incubed IT GmbH, Infineon Technologies Austria AG, Isovolta AG, Kostwein Holding GmbH, and Privatstiftung Kärntner Sparkasse.
We are greatful to the anonymous reviewers for their helpful comments.

\bibliographystyle{splncs04}
\bibliography{reference}

\begin{thebibliography}{10}
\providecommand{\url}[1]{\texttt{#1}}
\providecommand{\urlprefix}{URL }
\providecommand{\doi}[1]{https://doi.org/#1}

\bibitem{abels2021train}
Abels, D., Jordi, J., Ostrowski, M., Schaub, T., Toletti, A., Wanko, P.: Train
  scheduling with hybrid {ASP}. Theory and Practice of Logic Programming
  \textbf{21}(3),  317--347 (2021). \doi{10.1017/S1471068420000046}

\bibitem{abseher2016shift}
Abseher, M., Gebser, M., Musliu, N., Schaub, T., Woltran, S.: Shift design with
  answer set programming. Fundamenta Informaticae  \textbf{147}(1),  1--25
  (2016). \doi{10.3233/FI-2016-1396}

\bibitem{ali2023flexible}
Ali, R., El-Kholany, M., Gebser, M.: Flexible job-shop scheduling for
  semiconductor manufacturing with hybrid answer set programming (application
  paper). In: Proceedings of the Twenty-fifth International Symposium on
  Practical Aspects of Declarative Languages (PADL'23). pp. 85--95. Springer
  (2023). \doi{10.1007/978-3-031-24841-2_6}

\bibitem{balduccini2011industrial}
Balduccini, M.: Industrial-size scheduling with {ASP}+{CP}. In: Proceedings of
  the Eleventh International Conference on Logic Programming and Nonmonotonic
  Reasoning (LPNMR'11). pp. 284--296. Springer (2011).
  \doi{10.1007/978-3-642-20895-9_33}

\bibitem{bainkaokscsotawa18a}
Banbara, M., Inoue, K., Kaufmann, B., Okimoto, T., Schaub, T., Soh, T., Tamura,
  N., Wanko, P.: teaspoon: Solving the curriculum-based course timetabling
  problems with answer set programming. Annals of Operations Research
  \textbf{275}(1),  3--37 (2019). \doi{10.1007/s10479-018-2757-7}

\bibitem{brusch90a}
Brucker, P., Schlie, R.: Job-shop scheduling with multi-purpose machines.
  Computing  \textbf{45}(4),  369--375 (1990). \doi{10.1007/BF02238804}

\bibitem{cafageiakakrlemarisc19a}
Calimeri, F., Faber, W., Gebser, M., Ianni, G., Kaminski, R., Krennwallner, T.,
  Leone, N., Maratea, M., Ricca, F., Schaub, T.: {ASP-Core-2} input language
  format. Theory and Practice of Logic Programming  \textbf{20}(2),  294--309
  (2020). \doi{10.1017/S1471068419000450}

\bibitem{ceylan2021coordinated}
Ceylan, Z., Tozan, H., Bulkan, S.: A coordinated scheduling problem for the
  supply chain in a flexible job shop machine environment. Operational Research
   \textbf{21},  875--900 (2021). \doi{10.1007/s12351-020-00615-0}

\bibitem{cotmal06a}
Cotton, S., Maler, O.: Fast and flexible difference constraint propagation for
  {DPLL(T)}. In: Proceedings of the Ninth International Conference on Theory
  and Applications of Satisfiability Testing (SAT'06). pp. 170--183. Springer
  (2006). \doi{10.1007/11814948_19}

\bibitem{da2019industrial}
Da~Col, G., Teppan, E.: Industrial size job shop scheduling tackled by present
  day {CP} solvers. In: Proceedings of the Twenty-fifth International
  Conference on Principles and Practice of Constraint Programming (CP'19). pp.
  144--160. Springer (2019). \doi{10.1007/978-3-030-30048-7_9}

\bibitem{dogagrmamopo21a}
Dodaro, C., Galat{\`{a}}, G., Grioni, A., Maratea, M., Mochi, M., Porro, I.: An
  {ASP}-based solution to the chemotherapy treatment scheduling problem. Theory
  and Practice of Logic Programming  \textbf{21}(6),  835--851 (2021).
  \doi{10.1017/S1471068421000363}

\bibitem{eiter2022answer}
Eiter, T., Geibinger, T., Musliu, N., Oetsch, J., Skocovsk{\'{y}}, P.,
  Stepanova, D.: Answer-set programming for lexicographical makespan
  optimisation in parallel machine scheduling. In: Proceedings of the
  Eighteenth International Conference on Principles of Knowledge Representation
  and Reasoning (KR'21). pp. 280--290. {AAAI} Press (2021).
  \doi{10.24963/kr.2021/27}

\bibitem{el2022problem}
El-Kholany, M., Gebser, M., Schekotihin, K.: Problem decomposition and
  multi-shot {ASP} solving for job-shop scheduling. Theory and Practice of
  Logic Programming  \textbf{22}(4),  623--639 (2022).
  \doi{10.1017/S1471068422000217}

\bibitem{ellis2004scheduling}
Ellis, K., Lu, Y., Bish, E.: Scheduling of wafer test processes in
  semiconductor manufacturing. International Journal of Production Research
  \textbf{42}(2),  215--242 (2004). \doi{10.1080/0020754031000118116}

\bibitem{francescutto2021solving}
Francescutto, G., Schekotihin, K., El-Kholany, M.: Solving a multi-resource
  partial-ordering flexible variant of the job-shop scheduling problem with
  hybrid {ASP}. In: Proceedings of the Seventeenth European Conference on
  Logics in Artificial Intelligence (JELIA'21). pp. 313--328. Springer (2021).
  \doi{10.1007/978-3-030-75775-5_21}

\bibitem{DBLP:journals/mor/GareyJS76}
Garey, M., Johnson, D., Sethi, R.: The complexity of flowshop and jobshop
  scheduling. Mathematics of Operations Research  \textbf{1}(2),  117--129
  (1976). \doi{10.1287/moor.1.2.117}

\bibitem{PotasscoUserGuide19}
Gebser, M., Kaminski, R., Kaufmann, B., Lindauer, M., Ostrowski, M., Romero,
  J., Schaub, T., Thiele, S., Wanko, P.: Potassco User Guide (2019),
  \url{http://potassco.org}

\bibitem{gekakasc17a}
Gebser, M., Kaminski, R., Kaufmann, B., Schaub, T.: Multi-shot {ASP} solving
  with clingo. Theory and Practice of Logic Programming  \textbf{19}(1),
  27--82 (2019). \doi{10.1017/S1471068418000054}

\bibitem{gebser2016theory}
Gebser, M., Kaminski, R., Kaufmann, B., Ostrowski, M., Schaub, T., Wanko, P.:
  Theory solving made easy with clingo 5. In: Technical Communications of the
  Thirty-second International Conference on Logic Programming (ICLP'16). pp.
  2:1--2:15. Schloss Dagstuhl (2016). \doi{10.4230/OASIcs.ICLP.2016.2}

\bibitem{gran2015mixed}
Gran, S., Ismail, I., Ajol, T., Ibrahim, A.: Mixed integer programming model
  for flexible job-shop scheduling problem ({FJSP}) to minimize makespan and
  total machining time. In: Proceedings of the International Conference on
  Computer, Communications, and Control Technology (I4CT). pp. 413--417. IEEE
  (2015). \doi{10.1109/I4CT.2015.7219609}

\bibitem{ham2021energy}
Ham, A., Park, M., Kim, K.: Energy-aware flexible job shop scheduling using
  mixed integer programming and constraint programming. Mathematical Problems
  in Engineering  \textbf{2021}(Article ID 8035806),  1--12 (2021).
  \doi{10.1155/2021/8035806}

\bibitem{hassanzadeh2016two}
Hassanzadeh, A., Rasti-Barzoki, M., Khosroshahi, H.: Two new meta-heuristics
  for a bi-objective supply chain scheduling problem in flow-shop environment.
  Applied Soft Computing  \textbf{49},  335--351 (2016).
  \doi{10.1016/j.asoc.2016.08.019}

\bibitem{janhunen2017clingo}
Janhunen, T., Kaminski, R., Ostrowski, M., Schellhorn, S., Wanko, P., Schaub,
  T.: Clingo goes linear constraints over reals and integers. Theory and
  Practice of Logic Programming  \textbf{17}(5-6),  872--888 (2017).
  \doi{10.1017/S1471068417000242}

\bibitem{kopp2020smt2020}
Kopp, D., Hassoun, M., Kalir, A., M{\"o}nch, L.: S{MT}2020—{A} semiconductor
  manufacturing testbed. IEEE Transactions on Semiconductor Manufacturing
  \textbf{33}(4),  522--531 (2020). \doi{10.1109/TSM.2020.3001933}

\bibitem{kotaalelgese22a}
Kov{\'{a}}cs, B., Tassel, P., Ali, R., El-Kholany, M., Gebser, M., Seidel, G.:
  A customizable simulator for artificial intelligence research to schedule
  semiconductor fabs. In: Proceedings of the Thirty-third Annual SEMI Advanced
  Semiconductor Manufacturing Conference (ASMC'22). pp. 106--111. IEEE (2022).
  \doi{10.1109/ASMC54647.2022.9792520}

\bibitem{lesliePandemicScramblesSemiconductor2022}
Leslie, M.: Pandemic scrambles the semiconductor supply chain. Engineering
  \textbf{9},  10--12 (2022). \doi{10.1016/j.eng.2021.12.006}

\bibitem{li2016effective}
Li, X., Gao, L.: An effective hybrid genetic algorithm and tabu search for
  flexible job shop scheduling problem. International Journal of Production
  Economics  \textbf{174},  93--110 (2016). \doi{10.1016/j.ijpe.2016.01.016}

\bibitem{lifschitz19a}
Lifschitz, V.: Answer Set Programming. Springer (2019).
  \doi{10.1007/978-3-030-24658-7}

\bibitem{monchSurveyProblemsSolution2011}
M{\"{o}}nch, L., Fowler, J., Dauz{\`{e}}re{-}P{\'{e}}r{\`{e}}s, S., Mason, S.,
  Rose, O.: A survey of problems, solution techniques, and future challenges in
  scheduling semiconductor manufacturing operations. Journal of Scheduling
  \textbf{14}(6),  583--599 (2011). \doi{10.1007/s10951-010-0222-9}

\bibitem{pfund2008multi}
Pfund, M., Balasubramanian, H., Fowler, J., Mason, S., Rose, O.: A
  multi-criteria approach for scheduling semiconductor wafer fabrication
  facilities. Journal of Scheduling  \textbf{11}(1),  29--47 (2008).
  \doi{10.1007/s10951-007-0049-1}

\bibitem{pfundSemiconductorManufacturingScheduling2006}
Pfund, M., Mason, S., Fowler, J.: Semiconductor manufacturing scheduling and
  dispatching. In: Handbook of {{Production Scheduling}}, pp. 213--241.
  {Springer} (2006). \doi{10.1007/0-387-33117-4_9}

\bibitem{rigralmaliiile12a}
Ricca, F., Grasso, G., Alviano, M., Manna, M., Lio, V., Iiritano, S., Leone,
  N.: Team-building with answer set programming in the {G}ioia-{T}auro seaport.
  Theory and Practice of Logic Programming  \textbf{12}(3),  361--381 (2012).
  \doi{10.1017/S147106841100007X}

\bibitem{sahraeian2017new}
Sahraeian, R., Rohaninejad, M., Fadavi, M.: A new model for integrated lot
  sizing and scheduling in flexible job shop problem. Journal of Industrial and
  Systems Engineering  \textbf{10}(3),  72--91 (2017),
  \url{https://www.jise.ir/article_44919.html}

\bibitem{taillard93a}
Taillard, E.: Benchmarks for basic scheduling problems. European Journal of
  Operational Research  \textbf{64}(2),  278--285 (1993).
  \doi{10.1016/0377-2217(93)90182-M}

\bibitem{tassel2021multi}
Tassel, P., Rbaia, M.: A multi-shot {ASP} encoding for the aircraft routing and
  maintenance planning problem. In: Proceedings of the Seventeenth European
  Conference on Logics in Artificial Intelligence (JELIA'21). pp. 442--457.
  Springer (2021). \doi{10.1007/978-3-030-75775-5_30}

\bibitem{upasani2006problem}
Upasani, A., Uzsoy, R., Sourirajan, K.: A problem reduction approach for
  scheduling semiconductor wafer fabrication facilities. IEEE Transactions on
  Semiconductor Manufacturing  \textbf{19}(2),  216--225 (2006).
  \doi{10.1109/TSM.2006.873510}

\bibitem{wang2001effective}
Wang, L., Zheng, D.: An effective hybrid optimization strategy for job-shop
  scheduling problems. Computers \& Operations Research  \textbf{28}(6),
  585--596 (2001). \doi{10.1016/S0305-0548(99)00137-9}

\bibitem{waschneckDeepReinforcementLearning2018}
Waschneck, B., Reichstaller, A., Belzner, L., Altenmüller, T., Bauernhansl,
  T., Knapp, A., Kyek, A.: Deep reinforcement learning for semiconductor
  production scheduling. In: Proceedings of the Twenty-ninth Annual SEMI
  Advanced Semiconductor Manufacturing Conference (ASMC'18). pp. 301--306. IEEE
  (2018). \doi{10.1109/ASMC.2018.8373191}

\bibitem{xing2010knowledge}
Xing, L., Chen, Y., Wang, P., Zhao, Q., Xiong, J.: A knowledge-based ant colony
  optimization for flexible job shop scheduling problems. Applied Soft
  Computing  \textbf{10}(3),  888--896 (2010). \doi{10.1016/j.asoc.2009.10.006}

\end{thebibliography}

\end{document}